\documentclass{article}

\usepackage[final,dblblindworkshop]{neurips_2026}
\usepackage[utf8]{inputenc}
\usepackage[T1]{fontenc}
\usepackage{booktabs}
\usepackage{graphicx}
\usepackage{microtype}
\usepackage{amsmath}
\usepackage{amssymb}
\usepackage{xcolor}
\usepackage{url}
\usepackage{float}
\IfFileExists{placeins.sty}{\usepackage{placeins}}{\newcommand{\FloatBarrier}{\clearpage}}

\graphicspath{{figures/}}

\workshoptitle{Physical Understanding for Decision-Making}
\title{Learning When to Trust and Reobserve Sparse-Frame VideoQA
from Temporal Sampling Perturbations}
\author{
Shujia Liu \\
Tsinghua Shenzhen International Graduate School, Tsinghua University \\
Shenzhen, 518055, China \\
\texttt{liusj25@mails.tsinghua.edu.cn}
}

\newcommand{\method}{\textsc{PhaseGuard}}
\newcommand{\methodplus}{\textsc{PhaseGuard+}}

\newcommand{\pp}{\,pp}

\begin{document}
\raggedbottom
\setlength{\intextsep}{8pt plus 1pt minus 1pt}
\maketitle

\begin{abstract}
Sparse-frame video-language models can change their answers when only the
sampling timestamps shift.  Raw agreement exposes this problem, but it need
not rank errors better than token confidence or identify which additional
observation will repair an answer.  We argue that answer risk and action value
must be estimated separately.  \methodplus{} probes a frozen VLM at controlled
temporal offsets, predicts the reliability of the phase-vote answer, and
predicts the example-wise correctness gain of a specific rescue action.  The
second score routes only examples expected to benefit from denser frames or a
stronger VLM.  Both predictors use lightweight models trained on public data;
the VLMs remain frozen.  We use 256 Perception Test videos for method
development and evaluate on a disjoint 256-video test set comprising 554
Physics questions.  For Qwen3-VL-8B, reliability AUROC rises from $68.0\%$ with
token softmax to $74.8\%$ ($+6.8\pp{}$, 95\% CI $[2.6,10.7]\pp{}$).  Routing 30\% of
questions to Qwen3-VL-32B reaches 67.87\% accuracy, compared with 63.66\% for
budget-matched random routing ($+4.21$ percentage points,
CI $[1.89,6.40]$).  All four tested rescue pairs significantly outperform
random routing.  Temporal sensitivity becomes actionable when the system
learns both whether an answer is unreliable
and whether an available action is likely to fix it.
\end{abstract}

\section{Introduction}

Video question answering (VideoQA) often depends on a short event, such as an
occluded object moving or an interaction changing state
~\citep{xiao2021nextqa,wu2021star,patraucean2023perception}.  Yet a
bandwidth- or latency-limited system may observe only a few frames
~\citep{mangalam2023egoschema,wu2024longvideobench,fu2025videomme}.  Before a
physical judgment is passed downstream, the system must decide whether to
accept it or request another observation.  A useful trigger must answer two
different questions: \emph{is the current answer unreliable?}, and \emph{will
this available observation action repair it?}

\begin{figure}[H]
  \centering
  \includegraphics[width=\linewidth]{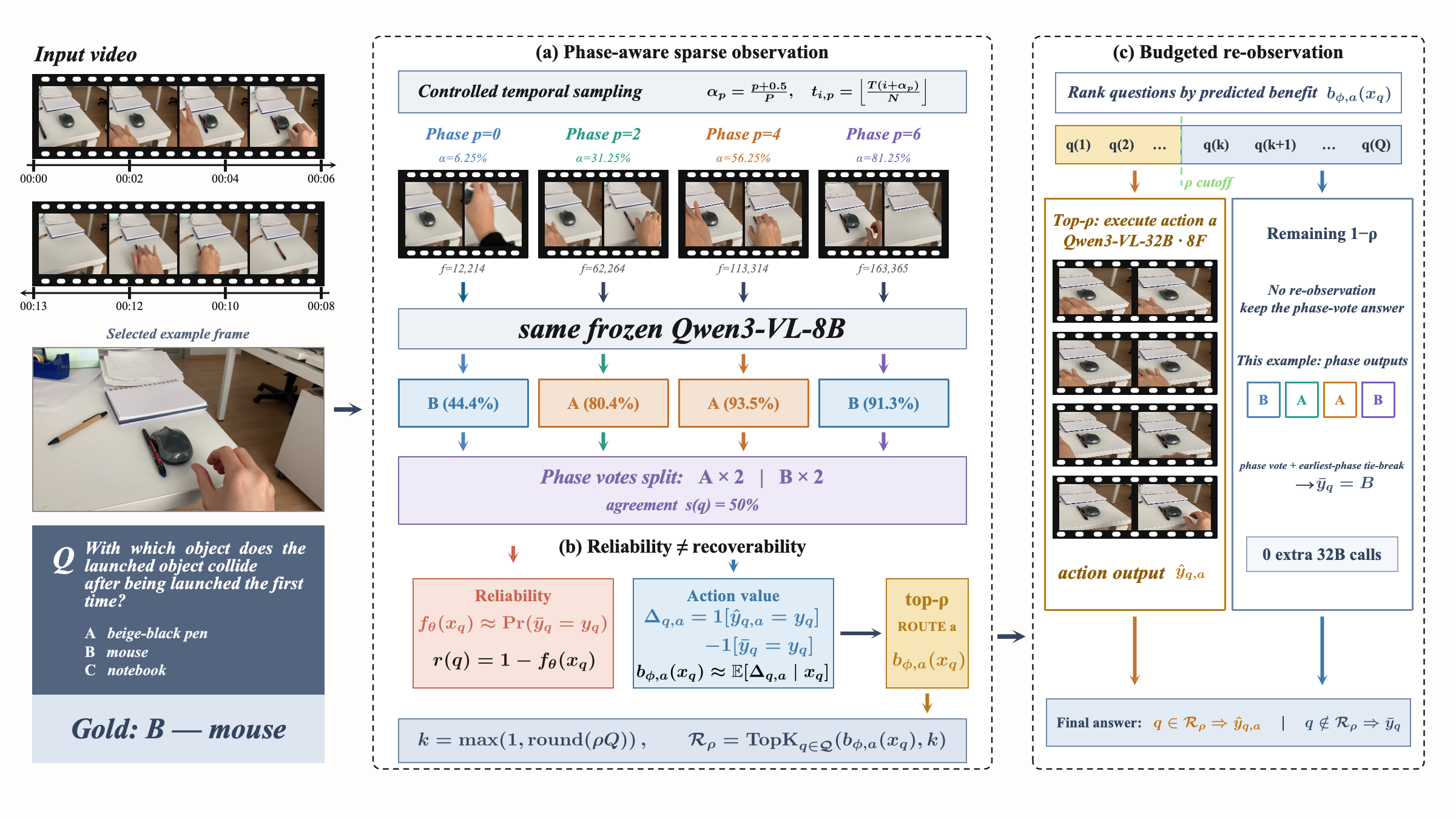}
  \caption{\textbf{Phase-aware selective re-observation.}
  \textbf{(a)} For a held-out Perception Test collision question, four
  controlled two-frame phases from the same frozen Qwen3-VL-8B yield
  $[B,A,A,B]$ ($A$: beige--black pen, $B$: mouse, $C$: notebook; gold $B$).
  The $2{:}2$ tie is resolved by the earliest-phase rule, giving
  $\bar y_q=B$ with phase consensus $s_q=50\%$.
  \textbf{(b)} The observable perturbation trace $x_q$ supports two separately
  supervised estimates: current-answer reliability $f_\theta(x_q)$ and the
  signed correctness gain $b_{\phi,a}(x_q)$ of a specified action $a$.
  \textbf{(c)} For a pool of $Q$ questions, the policy ranks predicted action
  benefits and re-observes only the top
  $K=\max(1,\operatorname{round}(\rho Q))$ questions with
  Qwen3-VL-32B/8F; all remaining questions retain their phase-vote answers.
  The displayed example follows the blue no-re-observation branch, while the
  orange branch visualizes the candidate action and its counterfactual output.}
  \label{fig:overview}
\end{figure}

Figure~\ref{fig:overview} summarizes the complete decision process.  Four
fixed-budget observations of the same video--question pair produce the response
trace $[B,A,A,B]$: the model, prompt, decoding, frame count, and temporal
coverage are unchanged, while only the sampling phase varies.  The trace is
then used to estimate two different quantities---whether the phase-vote answer
is correct and whether a specified re-observation action is expected to improve
correctness.  Under a fixed rescue budget, questions are ranked by the second
score and only the top-ranked subset is re-observed.

Frame selectors and token-compression methods optimize which visual evidence to
process~\citep{tang2025aks,buch2025flexible,shen2025longvu,suo2025trial};
selective VQA estimates whether an answer should be withheld
~\citep{whitehead2022reliable,dancette2023peers,khan2024consistency}; and
active-acquisition and routing methods allocate costly observations or stronger
predictors~\citep{shim2018active,nan2017adaptive,mozannar2020defer,
dekoninck2025routing}.  These objectives are not interchangeable.  In sparse
VideoQA, a likely error is not necessarily repairable by a particular action,
and an unnecessary action can overturn an already-correct answer.  \methodplus{}
therefore turns temporal perturbations into a decision signal for both trusting
the current answer and choosing whether a concrete rescue action is worthwhile.

We make three contributions.  (i) We formulate \emph{selective re-observation}
for sparse VideoQA: deciding whether to accept a sparse answer or invoke a
candidate observation action whose effect may be beneficial or harmful.  (ii)
We introduce \methodplus{}, which probes a frozen VLM at controlled temporal
offsets and separately predicts current-answer correctness and the signed
correctness gain of the candidate action.  A top-$K$ policy ranks questions by
the latter score and allocates a fixed re-observation budget without updating
the VLM.  (iii) On a disjoint public test set spanning multiple base
VLMs and rescue actions, \methodplus{} raises Qwen3-VL-8B correctness-ranking
AUROC from $68.0\%$ to $74.8\%$ ($+6.8\pp{}$) and outperforms budget-matched
random routing by $4.21\pp{}$ at a 30\% rescue rate; all four base--rescue pairs
yield significant gains.

\section{Related work}

\paragraph{Physical and temporal reasoning from video.}
VideoQA benchmarks cover causal events, situated interactions, physical
properties, and long-context judgments
~\citep{xiao2021nextqa,wu2021star,patraucean2023perception,
mangalam2023egoschema,wu2024longvideobench,fu2025videomme,
fang2024mmbenchvideo}.  Newer diagnostics more directly test frame order,
reasoning across separated moments, continuous perception, and prerequisites
for physical action~\citep{shangguan2025tomato,chen2024rextime,xun2025rtvbench,
gundawar2025pacbench}.  They consistently find substantial gaps even in strong
multimodal models.  Related physical benchmarks study grounded reachability,
physical commonsense, and conservation under transformation
~\citep{zhou2025physvlm,meng2025phygen,luo2026conservation}.  We complement
these task-level evaluations with an example-level question: whether the sparse
evidence supporting one particular physical answer is stable enough to use.

\paragraph{Long-video representation and temporal modeling.}
Time-aware instruction tuning, temporal attention, streaming memory, and
spatiotemporal compression increase the amount or quality of visual context
available to a Video-LLM~\citep{ren2024timechat,tan2024koala,
qian2024streaming,nie2024slowfocus,xu2024slotvlm,shen2025longvu,
islam2025bimba,rasekh2025stacked}.  Other methods learn question-conditioned
representations or improve temporal alignment and grounded reasoning
~\citep{zou2024videodistill,huang2024vtimellm,sarkar2025selfalignment,
xiao2024grounded}.  These approaches aim to improve the underlying predictor.
Our VLMs remain frozen: the object of study is the reliability of a sparse
prediction and the value of a subsequent observation action.

\paragraph{Frame selection and adaptive observation.}
Adaptive frame selection has been studied for efficient recognition and video
reasoning~\citep{wu2019adaframe,buch2025flexible,tang2025aks,
hu2025frameselection,diko2025rewind}.  Recent training-free selectors retrieve
query-relevant and diverse frames or adapt their resolution
~\citep{zhang2025qframe,sun2025mdp3}; iterative inference can also curate
context with the Video-LLM itself~\citep{arnab2025temporalcot}.  Sample scaling
instead aggregates predictions from multiple visual contexts
~\citep{suo2025trial}, and diagnostics confirm that apparently innocuous
sampling choices change VLM accuracy~\citep{brkic2025sampling,
yu2026virtuebench}.  \method{} does not claim to select the best frames.  It
deliberately applies a small, predefined family of timestamp changes and treats
the resulting response trace as evidence about the current answer.

\paragraph{Active acquisition, deferral, and routing.}
Active feature acquisition learns when additional evidence is worth its test-time
cost~\citep{shim2018active,nan2017adaptive}, while learning-to-defer and model
routing choose between predictors with different capabilities and costs
~\citep{mozannar2020defer,dekoninck2025routing,chuang2025selfref,
feng2025graphrouter,shirkavand2025cscr,dann2025routing}.  VLM selection further
shows that capability depends on the input domain rather than a single global
model ranking~\citep{yi2024swab}.  Most gates nevertheless estimate difficulty,
confidence, or expected response quality.  Our acquisition target is the
\emph{change} in correctness caused by a specified video observation action
relative to the current sparse answer.  The action can therefore be beneficial,
neutral, or harmful on different examples.

\paragraph{Selective prediction and VQA reliability.}
Selective classification trades coverage for risk
~\citep{geifman2017selective,geifman2019selectivenet}, and recent failure
detection work continues to address overconfident errors
~\citep{liu2024typicalness}.  Reliable VQA extends selective prediction to
multimodal inputs and learns selection functions from labeled data
~\citep{whitehead2022reliable,dancette2023peers}.  Other work estimates VLM
reliability from rephrasing consistency, learned multimodal uncertainty,
explicit abstention tokens, or calibrated perturbation tests
~\citep{khan2024consistency,ding2025lvlms,schmalfuss2025parc,
lafon2025vilu,cohen2024idk}.  Recent audits also expose confident multimodal
failures and sensitivity to modality perturbations
~\citep{liu2025ignorance,chowdhury2025avtrust}.  Our reliability score is
learned from a frozen model's response trace under temporal evidence
perturbations.  More importantly, reliability is not reused as a surrogate for
recovery: a separate action-conditioned target is learned for routing.

\section{\methodplus{}}

\subsection{Overview and problem formulation}

Let $V=(v_0,\ldots,v_{T-1})$ be a video, $q$ a multiple-choice question, and
$\mathcal Y_q$ its answer set.  A frozen Video-LLM $F_m$ receives an observation
$O\subset V$ and returns both an answer $\hat y\in\mathcal Y_q$ and an option
distribution $\boldsymbol\pi\in\Delta^{|\mathcal Y_q|-1}$.  The initial system
is deliberately observation-limited: it may show only $N=2$ frames to the
model.  After obtaining a sparse answer, it must either accept that answer or
invoke a candidate rescue action $a$, such as evaluating eight frames with a
larger VLM.

\methodplus{} separates this decision into two predictions.  A reliability
model estimates whether the current sparse answer is correct; an
action-specific benefit model estimates whether invoking $a$ will change
correctness for this example.  Both models use the same temporal-perturbation
trace and are lightweight; all Video-LLMs remain frozen.  This distinction is
central: an unreliable answer need not be repairable by $a$, while invoking $a$
can replace an already-correct answer with an incorrect one.

Figure~\ref{fig:overview} gives the end-to-end pipeline.  The following
subsections formalize its three stages: phase-aware sparse observation, the
perturbation-trace representation and dual-score estimators, and budgeted
re-observation.

\subsection{Temporal phase probing}

\paragraph{Phase construction.}
For a video--question pair, \method{} asks whether the answer changes when only
the sampling timestamps shift.  It evaluates deterministic sparse observations
that differ in temporal phase; the model, prompt, decoding, frame count, and
temporal coverage remain fixed.  Dividing the timeline into $N$ equal bins,
phase $p$ on a grid of $P=8$ selects

\begin{equation}
  \alpha_p=\frac{p+0.5}{P}, \qquad
  t_{i,p}=\left\lfloor\frac{T(i+\alpha_p)}{N}\right\rfloor,
  \quad i=0,\ldots,N-1,
\end{equation}
clipped to valid indices.  We use $N=2$ and
$\mathcal P=\{0,2,4,6\}$.  Every probe spans the full video with the same frame
count and ordering; only the fractional position inside each half changes.  In
particular, a phase shift neither enlarges the temporal field of view nor gives
the model more frames.  It therefore isolates sensitivity to which instant is
observed within each temporal bin, rather than conflating sensitivity with a
change in input budget.

\paragraph{Sparse answer and agreement baseline.}
For question $q$, probe $p$ returns $(\hat y_{q,p},\boldsymbol\pi_{q,p})$.
The voted answer and phase-consensus score are
\begin{equation}
  \bar y_q=\arg\max_y\sum_{p\in\mathcal P}\mathbf 1[\hat y_{q,p}=y],
  \qquad
  s_q=\frac{1}{|\mathcal P|}\max_y\sum_{p\in\mathcal P}
  \mathbf 1[\hat y_{q,p}=y].
\end{equation}
Ties use the earliest probe.  Following selective prediction notation
~\citep{geifman2017selective,whitehead2022reliable}, the selection function
$g_q$ and selective output $h_q$ are
\begin{equation}
  g_q=\mathbf 1[s_q=1], \qquad
  h_q=\begin{cases}\bar y_q,&g_q=1,\\ \bot,&g_q=0,\end{cases}
\end{equation}
where $\bot$ denotes abstention.  Unanimity does not guarantee correctness.  We
also evaluate $s_q$ as a ranking score to distinguish the selection rule from
general confidence estimation.

\subsection{Perturbation trace representation}

The four probes define a deployment-observable trace
$\mathcal T_q=\{(\hat y_{q,p},\boldsymbol\pi_{q,p}):p\in\mathcal P\}$.
We map it to a fixed-dimensional vector $x_q$ using four groups of statistics.
First, \emph{vote features} record consensus, vote margin, the number of answer
switches, and entropy of the empirical vote histogram.  Second,
\emph{confidence features} summarize, over phases, the probability assigned to
$\bar y_q$, the maximum option probability, the top-two margin, and predictive
entropy using their mean, extrema, range, and variance.  Third,
\emph{shift features} measure how the entire option distribution changes.  For
example, with $\bar{\boldsymbol\pi}_q=|\mathcal P|^{-1}
\sum_p\boldsymbol\pi_{q,p}$, we include
\begin{equation}
 d_{\mathrm{JS}}(q)=\frac{1}{|\mathcal P|}\sum_{p\in\mathcal P}
 D_{\mathrm{KL}}\!\left(\boldsymbol\pi_{q,p}\,\middle\|\,
 \bar{\boldsymbol\pi}_q\right),
\end{equation}
along with mean and maximum pairwise $\ell_1$ distance.  Fourth, directly
observable question statistics include the number of options and question and
option lengths.

The \emph{single-model} variant uses only the target VLM trace.  The optional
\emph{full} variant additionally summarizes the other two open VLMs' votes,
confidence, and agreement with the target model.  It never exposes their literal
answer indices to the estimator.  Gold answers, Perception Test reasoning tags,
and other benchmark-only annotations are excluded from both variants.  This
prevents the learned models from exploiting features unavailable for a new
question.

\subsection{Dual-score estimation}

\subsubsection{Reliability scoring}

Define the correctness label $z_q=\mathbf 1[\bar y_q=y_q]$.  A classifier
$f_\theta(x_q)$ is trained to estimate $\Pr(z_q=1\mid x_q)$.  Token confidence
remains a useful complementary signal, so model selection may rank-blend the
learned score with the mean probability assigned to $\bar y_q$.  Let
$F_{\mathrm{dev}}(\cdot)$ denote the empirical percentile map fixed on the
development pool.  The final reliability score is
\begin{equation}
 R_q=(1-\lambda)F_{\mathrm{dev}}(c_q)
   +\lambda F_{\mathrm{dev}}(f_\theta(x_q)),
 \qquad r_q=1-R_q,
\end{equation}
where $c_q$ is token confidence, $R_q$ ranks answer correctness, and $r_q$ is
error risk.  Percentile mapping makes heterogeneous score scales comparable
without using test labels.  We select the estimator, feature scope, and
$\lambda\in\{0,.05,\ldots,1\}$ by five-fold, video-grouped out-of-fold AUROC.
Candidate classifiers are regularized logistic regression, Extra Trees, random
forests, and histogram gradient boosting.  After selection, the chosen model is
refit on all development videos and frozen.

\subsubsection{Action-value estimation}

Reliability answers whether the current prediction is likely to be correct; it
does not specify what an additional observation will do.  For a candidate rescue
action $a$---for example, eight frames with a 32B model---we define its realized
signed benefit
\begin{equation}
  \Delta_{q,a}=\mathbf 1[\hat y_{q,a}=y_q]
  -\mathbf 1[\bar y_q=y_q] \in \{-1,0,1\}.
\end{equation}
The three values have operational meaning: $+1$ means the action repairs an
error, $0$ means it leaves correctness unchanged, and $-1$ means it breaks a
correct answer.  For each base--action pair, a regressor
$b_{\phi,a}(x_q)$ predicts $\mathbb E[\Delta_{q,a}\mid x_q]$.  We consider
ridge, Extra Trees, and random-forest regressors.  Their hyperparameters,
feature scope, and routing rate are selected only by video-grouped development
predictions, using gain over matched random routing as the criterion.

\subsection{Budgeted re-observation policy}

For a batch of $M$ questions and a rescue budget $\rho$, the policy may invoke
$a$ on at most $K=\lfloor\rho M\rceil$ examples.  Because each invocation has
the same cost within an evaluated action pair, maximizing predicted total gain
reduces to
\begin{equation}
 S_{a,\rho}=\underset{S\subseteq\{1,\ldots,M\},\ |S|=K}{\arg\max}
 \sum_{q\in S} b_{\phi,a}(x_q),
 \qquad
 \tilde y_q=\begin{cases}
   \hat y_{q,a}, & q\in S_{a,\rho},\\
   \bar y_q, & \text{otherwise}.
 \end{cases}
\end{equation}
Thus the router simply selects the top $K$ predicted benefits.  Its realized
accuracy equals the base majority accuracy plus
$M^{-1}\sum_{q\in S_{a,\rho}}\Delta_{q,a}$.  In contrast, routing by
disagreement or error risk prioritizes likely failures even when the specified
action cannot repair them.  Matched random routing draws the same number $K$ of
examples, isolating the value of the learned ordering from the cost of making
additional calls.

\paragraph{Inference cost.}
The default trace requires four calls to the target frozen VLM, each with two
frames; all four phases can be batched.  The single-model variant adds no other
model calls before routing.  The full variant adds the corresponding trace
summaries from two open VLMs, while the rescue VLM is invoked only on the routed
fraction.  Section~4 reports a two-probe ablation and a one-call eight-frame
control so that reliability gains are not presented without their observation
cost.

\section{Experimental setup}

\paragraph{Data and split protocol.}
We use public CC BY real-world videos from Perception
Test~\citep{patraucean2023perception}.
The development pool contains 256 videos and 989 questions (514 Physics, 475
Semantics).  Reliability models use both categories because correctness is the
training target; recovery models use the 514 Physics questions that match the
downstream task.  Feature sets, estimator families, blend weights, rescue
actions, and routing rates are selected by five-fold video-grouped out-of-fold
evaluation and serialized before confirmatory evaluation.  The disjoint test
set contains another 256 videos with 554 Physics and 432 Semantics questions.
No video occurs in both pools, and no test label is used for model or
hyperparameter selection.  Physics is the only confirmatory target below;
Semantics is retained for cross-category development diagnostics.

\paragraph{Models and inference.}
Models are Qwen3-VL-4B/8B-Instruct~\citep{bai2025qwen3vl} and
InternVL3.5-8B-HF~\citep{wang2025internvl35}; Qwen3-VL-32B is rescue-only.
Chronological frames and a fixed zero-shot prompt produce one option letter via
greedy decoding.  Confidence is computed by aggregating first-step logits over
valid tokenizations of each option letter and normalizing across legal options.
The base answer is the four-phase majority.  Candidate rescue actions are
Qwen3-VL-32B with eight frames, or, for InternVL, the same 8B model with 16
frames.  The confirmatory sweep contains 12,940 model outputs with no missing
records, inference errors, or parse failures.

\paragraph{Protocol selection.}
For reliability, the development search compares the classifier families in
Section~3.4, single/full feature scopes, and 21 blend weights.  For recovery,
each base--action pair compares ridge, Extra Trees, and random forests at
$\rho\in\{.10,.15,.20,.25,.30\}$.  Selection uses only grouped out-of-fold
predictions.  The primary protocol, chosen by the largest development AUROC
gain and then the best action for that same base model, is Qwen3-VL-8B with a
random-forest reliability model ($\lambda=.95$) and an Extra-Trees benefit
model that routes 30\% to Qwen3-VL-32B/8F.  Results for every other locked
base--action pair are also reported.

\paragraph{Baselines and metrics.}
Reliability baselines are raw phase consensus and token softmax; we report
correctness-ranking AUROC, the risk--coverage curve, AURC, and accuracy at fixed
coverage.  Routing baselines include matched random, phase disagreement, low
softmax, predicted answer risk, always rescue, and an oracle that ranks the
realized $\Delta_{q,a}$.  The primary endpoints fixed before test aggregation
are Qwen-8B AUROC versus token softmax and Qwen-8B $\rightarrow$ Qwen-32B/8F
accuracy versus 30\% matched-random routing.  All confidence intervals use
10,000 bootstrap resamples of videos, preserving dependence among questions
from the same video.

\section{Results}

\subsection{Timestamp shifts expose a repeatable failure mode}

\begin{table}[H]
  \centering
  \small
  \setlength{\tabcolsep}{6.0pt}
  \caption{Fixed-budget temporal sensitivity on the 554-question Physics test
  set.  \emph{Disagree} is the fraction for which the four two-frame probes do
  not produce one unanimous answer.  \emph{Gain} is unanimous-subset accuracy
  minus full-coverage majority accuracy.}
  \label{tab:sensitivity}
  \begin{tabular}{lccccc}
    \toprule
    Model & Disagree & Unanimous coverage & Majority acc. & Unanimous acc. & Gain \\
    \midrule
    Qwen3-VL-4B      & 36.5 & 63.5 & 61.6 & 70.2 & $+8.6$ \\
    Qwen3-VL-8B      & 39.2 & 60.8 & 58.8 & 65.0 & $+6.1$ \\
    InternVL3.5-8B   & 35.2 & 64.8 & 60.6 & 69.6 & $+9.0$ \\
    \bottomrule
  \end{tabular}
\end{table}

Changing only the within-bin timestamps alters at least one answer for
$35.2$--$39.2\%$ of Physics questions (Table~\ref{tab:sensitivity}).  Raw
unanimity therefore identifies a safer subset: retained accuracy rises by
$6.1$--$9.0\pp{}$, and all three video-clustered 95\% intervals exclude zero
([2.8,9.5], [5.6,11.8], and [5.9,12.2]\pp{} for Qwen-8B, Qwen-4B, and
InternVL, respectively).  This operating point, however, discards more than a
third of the questions and does not rank the non-unanimous cases.

\begin{figure*}[t]
  \centering
  \includegraphics[width=.98\textwidth]{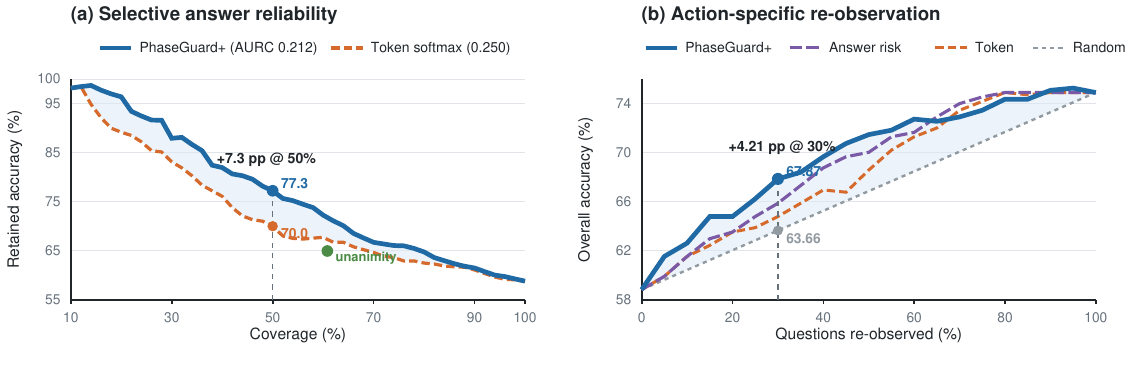}
  \caption{\textbf{Reliability ranking and action-specific re-observation on
  554 held-out Physics questions.} \textbf{(a)} Lower AURC denotes a better
  reliability ranking.  At 50\% coverage, \methodplus{} improves retained
  accuracy from 70.0\% to 77.3\%; the green marker denotes raw unanimity.
  \textbf{(b)} Policies rank questions for the same Qwen3-VL-32B/8F action.
  At the locked 30\% budget, action-value routing reaches 67.87\%, versus
  63.66\% for matched random.}
  \label{fig:main-results}
\end{figure*}

\subsection{Learning reliability from the perturbation trace}

Figure~\ref{fig:main-results}(a) evaluates whether the learned score places
correct answers ahead of errors.  At coverage $c$, only the top-$c$ fraction is
retained; at 100\%, ranking no longer matters and the curves meet.  For
Qwen-8B, correctness-ranking AUROC rises from $68.0\%$ to $74.8\%$
($+6.8\pp{}$, CI $[2.6,10.7]\pp{}$), close to the locked training OOF gain of
$6.5\pp{}$.  Five of six configurations have intervals above zero.  The
single-model variant still gains $4.0\pp{}$ ($[1.0,7.1]\pp{}$), while InternVL
gains $3.4\pp{}$ ($[1.5,5.4]\pp{}$).

\begin{table}[H]
  \centering
  \small
  \setlength{\tabcolsep}{7.0pt}
  \caption{Correctness-ranking AUROC (\%) on the confirmatory Physics set.
  \emph{Agreement} uses raw phase consensus; \emph{single} learns only from
  the target VLM trace; \emph{full} additionally uses cross-model summaries.}
  \label{tab:risk-ablation}
  \begin{tabular}{lcccc}
    \toprule
    Model & Agreement & Token softmax & \methodplus{} single & \methodplus{} full \\
    \midrule
    Qwen3-VL-4B      & 62.1 & 70.7 & 72.8 & \textbf{73.7} \\
    Qwen3-VL-8B      & 58.2 & 68.0 & 72.0 & \textbf{74.8} \\
    InternVL3.5-8B   & 62.5 & 73.1 & 74.2 & \textbf{76.5} \\
    \bottomrule
  \end{tabular}
\end{table}

Table~\ref{tab:risk-ablation} separates the source of the gain.  Raw agreement
is substantially weaker than token confidence, whereas a predictor learned
from the target model's trace improves all three models.  Adding cross-model
summaries gives the largest Qwen-8B gain.  Five of the six single/full
improvements over softmax have 95\% intervals above zero; the exception is the
Qwen-4B full variant ($+3.0\pp{}$, $[-0.6,6.6]\pp{}$).

The improvement also holds under selective prediction.  At 50\% coverage, the
learned score retains a subset that is $7.3\pp{}$ more accurate than the subset
chosen by token softmax.  Raw unanimity gives only one point (60.8\% coverage,
65.0\% accuracy), whereas the learned ranking supports any desired coverage.

\subsection{Action value targets recovery rather than uncertainty}

Figure~\ref{fig:main-results}(b) evaluates a different decision.  For a rescue
budget $\rho$, the top-$\rho$ fraction is sent to Qwen3-VL-32B/8F and the rest
keep the Qwen3-VL-8B answer.  Policies share the endpoints at $\rho=0$ and 1,
but need not improve monotonically because rescue may either fix or break an
answer.  We therefore rank signed action gain rather than answer uncertainty.

\begin{table}[H]
  \centering
  \small
  \setlength{\tabcolsep}{8.0pt}
  \caption{Routing the Qwen3-VL-8B majority answer to Qwen3-VL-32B/8F.
  Except for the no-rescue and always-rescue rows, all policies route exactly
  30\% of questions.  Phase ties are averaged uniformly.}
  \label{tab:primary-routing}
  \begin{tabular}{lcc}
    \toprule
    Routing policy & Rescue rate & Overall accuracy \\
    \midrule
    No rescue                    & 0\%   & 58.84 \\
    Matched random               & 30\%  & 63.66 \\
    Phase disagreement           & 30\%  & 64.56 \\
    Low token softmax            & 30\%  & 64.80 \\
    High predicted answer risk   & 30\%  & 65.88 \\
    High predicted action value  & 30\%  & \textbf{67.87} \\
    Always rescue                & 100\% & 74.91 \\
    Oracle action value          & 30\%  & 84.84 \\
    \bottomrule
  \end{tabular}
\end{table}

\begin{figure*}[t]
  \centering
  \includegraphics[width=\textwidth]{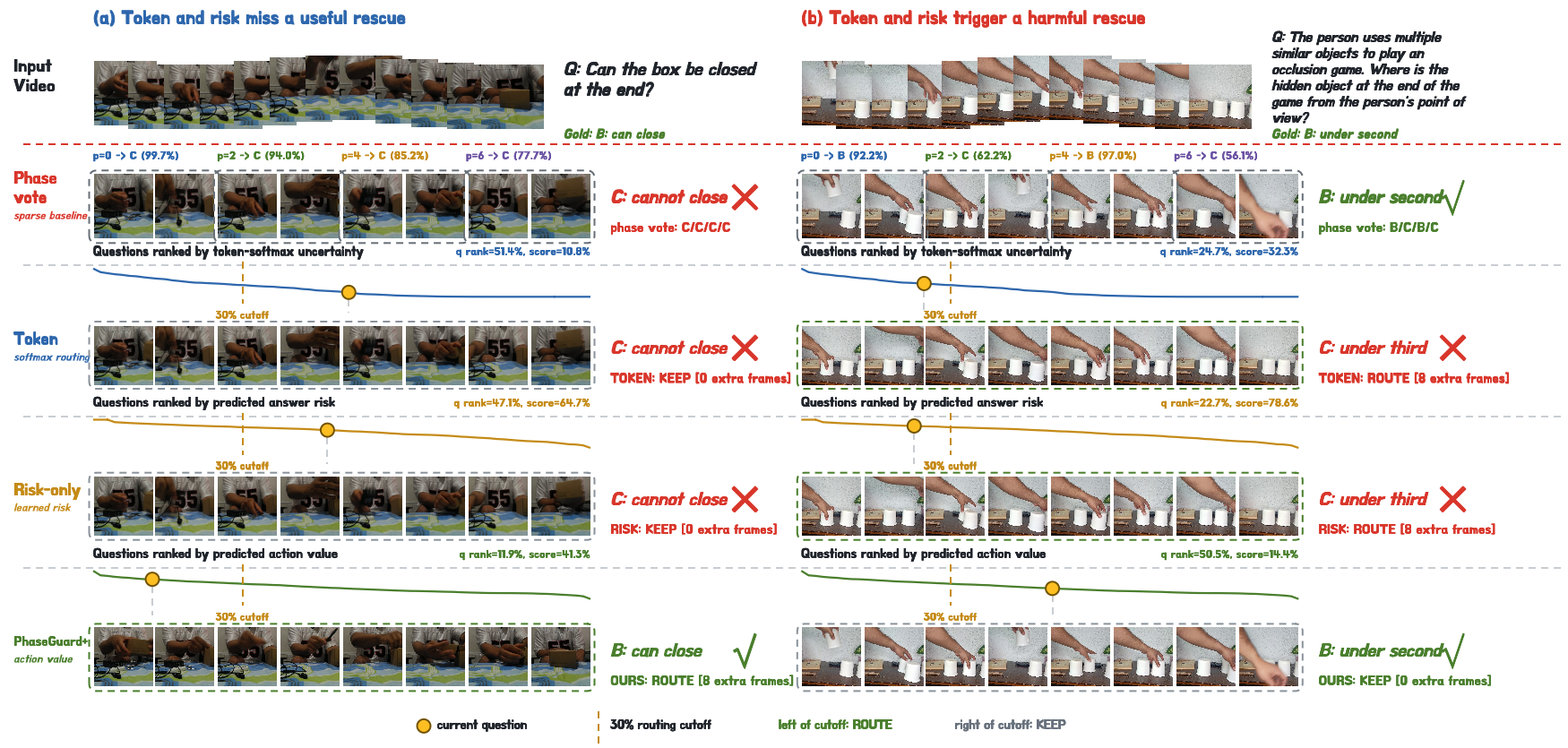}
  \caption{\textbf{Why answer uncertainty is not action value.}
  Two held-out test examples are shown under the same 30\% re-observation
  budget.  Each curve ranks all test questions by the corresponding routing
  score; the yellow marker locates the displayed question and the vertical
  dashed line denotes the routing cutoff.  \textbf{Left:} token-softmax and
  predicted answer risk keep an incorrect phase-vote answer, whereas
  \methodplus{} predicts positive benefit, invokes Qwen3-VL-32B/8F, and
  corrects the answer.  \textbf{Right:} token-softmax and answer risk route an
  initially correct answer, but the rescue action changes it to an incorrect
  one; \methodplus{} predicts low benefit and retains the correct phase-vote
  answer.  The examples illustrate that the probability of current error and
  the signed gain of a particular rescue action are different quantities.}
  \label{fig:qualitative-routing}
\end{figure*}

At the locked 30\% budget, action-value routing improves over phase
disagreement by $3.31\pp{}$ (95\% CI $[0.50,6.01]\pp{}$) and over matched
random routing by $4.21\pp{}$ ($[1.89,6.40]\pp{}$).  Its point estimate is also
higher than softmax routing by $3.07\pp{}$ and answer-risk routing by
$1.99\pp{}$, although the corresponding intervals cross zero.  Thus the
confirmatory evidence establishes an advantage over random allocation and raw
phase disagreement, but does not establish universal superiority over every
uncertainty router.  The 84.84\% oracle value shows that substantial headroom
remains in predicting action-specific benefit.  Across 10\%--50\% rescue,
action-value routing reaches 62.64\%--71.48\% accuracy versus
60.44\%--66.88\% for matched random.  Figure~\ref{fig:qualitative-routing}
shows both sides of this distinction: uncertainty can miss a beneficial rescue
or trigger a rescue that overturns an already-correct answer.

\begin{table}[H]
  \centering
  \small
  \setlength{\tabcolsep}{4.0pt}
  \caption{Action-specific recovery on 554 Physics questions.  $\Delta$ is
  learned minus matched-random accuracy with a video-clustered 95\% CI.}
  \label{tab:recovery}
  \begin{tabular}{lccccc}
    \toprule
    Base $\rightarrow$ rescue & Route & Base & Random & Learned & $\Delta$ (pp) \\
    \midrule
    Qwen-8B $\rightarrow$ Qwen-32B/8F & 30\% & 58.84 & 63.66 & \textbf{67.87} & $+4.21$ $[1.89,6.40]$ \\
    Qwen-4B $\rightarrow$ Qwen-32B/8F & 30\% & 61.55 & 65.55 & 68.59 & $+3.04$ $[0.81,5.09]$ \\
    IVL-8B $\rightarrow$ Qwen-32B/8F  & 25\% & 60.65 & 64.20 & 67.33 & $+3.13$ $[0.71,5.74]$ \\
    IVL-8B $\rightarrow$ IVL-8B/16F   & 30\% & 60.65 & 59.89 & 62.45 & $+2.56$ $[0.55,4.66]$ \\
    \bottomrule
  \end{tabular}
\end{table}

The primary Qwen-8B policy gains 9.03\pp{} over no rescue and 4.21\pp{} over
matched random; all four action pairs have intervals above zero
(Table~\ref{tab:recovery}).  Although InternVL/16F is harmful when always used
(58.12\% versus 60.65\%), selective routing reaches 62.45\%---a distinction
answer risk alone cannot make.

\subsection{Observation-budget control}

\begin{table}[H]
  \centering
  \small
  \setlength{\tabcolsep}{8.0pt}
  \caption{Qwen3-VL-8B observation controls on 554 Physics questions.  Four
  two-frame probes and one eight-frame call both sample eight frames.}
  \label{tab:observation-control}
  \begin{tabular}{lccc}
    \toprule
    Base observation & VLM calls & Total sampled frames & Accuracy \\
    \midrule
    One two-frame phase       & 1 & 2 & 57.04 \\
    Four two-frame phases, majority & 4 & 8 & 58.84 \\
    One eight-frame input     & 1 & 8 & \textbf{60.47} \\
    \bottomrule
  \end{tabular}
\end{table}

A single eight-frame input is $1.62\pp{}$ more accurate than the four-probe
majority, but the paired interval crosses zero ($[-1.83,5.11]\pp{}$).  Thus the
trace is not an eight-frame accuracy substitute: it exposes sample-specific
instability at the cost of four frozen-VLM calls, which can be batched.

\begin{table}[H]
  \centering
  \small
  \setlength{\tabcolsep}{5.5pt}
  \caption{Probe-count ablation for single-model Qwen3-VL-8B.  Estimators are
  selected on development; recovery routes 30\% to Qwen3-VL-32B/8F.
  \emph{Gain} is learned minus matched-random accuracy.}
  \label{tab:probe-ablation}
  \begin{tabular}{lccccc}
    \toprule
    Probes (phases) & Risk AUROC & Base acc. & Random & Learned & Gain \\
    \midrule
    1 (0)       & 71.0 & 57.22 & 62.52 & 64.62 & $+2.10$ \\
    2 (0,4)     & \textbf{72.9} & 57.22 & 62.52 & 66.79 & \textbf{$+4.27$} \\
    3 (0,2,4)   & 71.4 & 58.84 & 63.66 & 66.61 & $+2.95$ \\
    4 (0,2,4,6) & 72.1 & 58.84 & 63.66 & \textbf{67.87} & $+4.21$ \\
    \bottomrule
  \end{tabular}
\end{table}

Two probes retain most recovery performance (66.79\% versus 67.87\%) while
halving target-VLM calls.  Results are non-monotone because the phase set changes
both the base vote and feature distribution; this is a practical cost comparison,
not evidence that two probes always dominate.

\FloatBarrier
\section{Conclusion}

We study when a sparse-frame VideoQA system should trust its current answer
and when it should acquire an additional observation.  Our results show that
temporal sampling perturbations reveal failures that are not captured by token
confidence alone.  Building on this observation, \methodplus{} separately
estimates answer reliability and the expected benefit of a candidate rescue
action.  This separation enables more accurate selective prediction and more
effective budgeted re-observation across multiple base--rescue model pairs.

\paragraph{Limitations.}
Our evaluation is currently limited to multiple-choice Physics questions from
Perception Test.  Extending the approach to open-ended VideoQA and broader video
domains remains future work.

\bibliographystyle{plainnat}
\bibliography{references}

\clearpage
\appendix

\section{Reproducibility details}

\subsection{Prompt, decoding, and frame construction}

All Video-LLMs receive the same system instruction:
\begin{quote}
\small
``You answer multiple-choice questions about chronologically ordered frames
sampled from a real video. Use the visual evidence and physical evolution
across frames. Return exactly one option letter and no other text.''
\end{quote}
The user message contains the question followed by lettered answer options and
the instruction \emph{Answer with exactly one letter}.  Frames are supplied in
chronological order.  We use bfloat16 inference, SDPA attention, greedy decoding
with at most 16 new tokens, and no sampling.  An output is parsed only when it
contains one unambiguous legal option letter; the confirmatory run contains no
parse failures.

For confidence, we collect the first-step logits of every valid tokenization of
each legal option letter and normalize across options.  The probability used by
the token-softmax baseline is the mean, over the four probes, assigned to the
phase-majority answer.  Base traces use $N=2$, $P=8$, and
$\mathcal P=\{0,2,4,6\}$.  Candidate 8- and 16-frame actions use the same
binning rule with phase $p=4$ ($\alpha_p=56.25\%$), fixed before confirmatory
evaluation.  Majority ties are resolved by the earliest probe.  VLM generation
uses seed 1234; estimator selection uses seed 20260820.

\subsection{Feature inventory and model selection}

\begin{table}[H]
  \centering
  \small
  \setlength{\tabcolsep}{5.0pt}
  \caption{Deployment-observable perturbation features.  Literal predicted
  option identities, gold answers, and Perception Test reasoning annotations
  are excluded.}
  \label{tab:appendix-features}
  \begin{tabular}{p{.19\linewidth}p{.73\linewidth}}
    \toprule
    Feature group & Statistics supplied to the lightweight estimator \\
    \midrule
    Question structure & Number of options; question length; mean option length \\
    Phase votes & Consensus; vote margin; vote entropy; adjacent switches; opposite-phase agreement \\
    Confidence & Mean, minimum, maximum, range, and variance of voted-option probability; maximum probability; top-two margin; predictive entropy \\
    Distribution shift & Entropy of the mean distribution; Jensen--Shannon divergence; mean and maximum pairwise $\ell_1$ distance \\
    Cross-model (full) & Cross-model consensus; target--cross-model agreement; other-model confidence and consensus summaries \\
    \bottomrule
  \end{tabular}
\end{table}

\begin{table}[H]
  \centering
  \footnotesize
  \setlength{\tabcolsep}{3.6pt}
  \caption{Development-only estimator search.  \emph{Leaf} denotes the minimum
  samples per leaf; RF and ET denote random forests and Extra Trees.}
  \label{tab:appendix-search}
  \begin{tabular}{p{.20\linewidth}p{.48\linewidth}p{.23\linewidth}}
    \toprule
    Component & Candidates & Locked Qwen-8B configuration \\
    \midrule
    Reliability & Logistic regression $C\in\{.03,.1\}$; ET (1,000 trees, leaf 8/16); RF (1,000 trees, leaf 10); histogram gradient boosting; blend $\lambda\in\{0,.05,\ldots,1\}$ & RF, leaf 10; full features; $\lambda=.95$ \\
    Action value & Ridge $\alpha\in\{1,10\}$; ET (800 trees, leaf 8/16); RF (800 trees, leaf 10); $\rho\in\{.10,.15,.20,.25,.30\}$ & ET, leaf 8; full features; $\rho=.30$ \\
    \bottomrule
  \end{tabular}
\end{table}

All choices use five-fold GroupKFold with video identity as the group.  Missing
features are median-imputed; linear models additionally standardize features.
Classifier tree models use balanced class weights.  After selection by grouped
out-of-fold AUROC or gain over matched random, the chosen estimator is refit on
all 256 development videos and applied unchanged to the disjoint test set.

\clearpage
\section{Additional recovery analysis}

\subsection{Budget curves across rescue actions}

\begin{figure}[H]
  \centering
  \includegraphics[width=.98\linewidth]{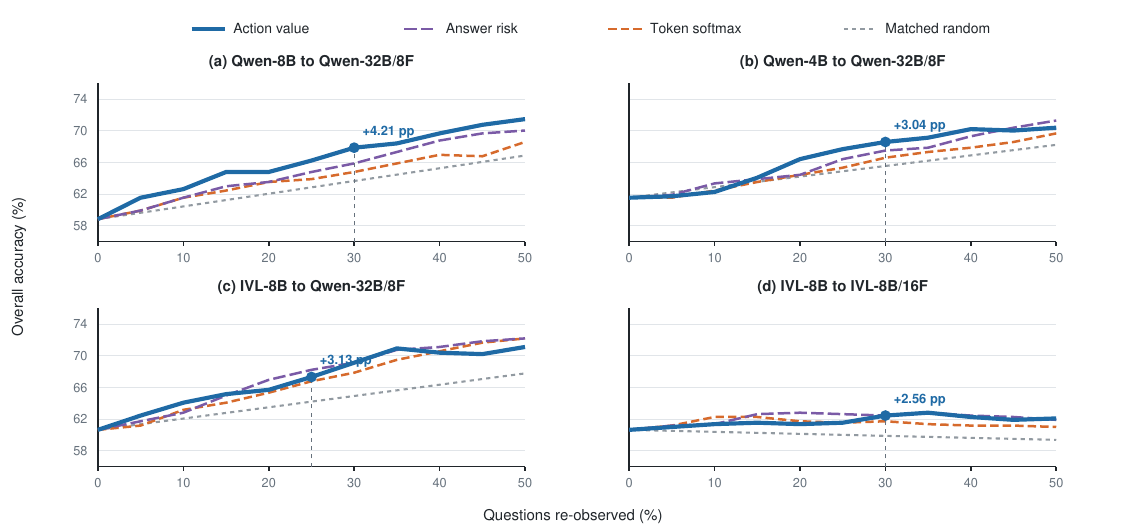}
  \caption{\textbf{Budgeted recovery across all four locked action pairs.}
  Each panel reuses one predictor frozen on the development set and varies only
  the fraction routed on the 554-question test set.  The blue marker denotes the
  development-selected rate; its annotation is the accuracy gain over expected
  matched-random routing at the same budget.  Action-value routing is most
  useful in the constrained-budget regime; the curves may cross at larger
  budgets because a rescue action can also overturn a correct answer.}
  \label{fig:appendix-budget-curves}
\end{figure}

\subsection{Repair--harm decomposition}

\begin{table}[H]
  \centering
  \small
  \setlength{\tabcolsep}{4.2pt}
  \caption{Outcome decomposition on the confirmatory test set.  \emph{All}
  columns use all 554 questions; \emph{routed} columns use only the selected
  subset at the locked rate.  Repair is wrong$\rightarrow$correct and harm is
  correct$\rightarrow$wrong.  Net gain is relative to no rescue over all 554
  questions.}
  \label{tab:appendix-transitions}
  \begin{tabular}{lccccc}
    \toprule
    Base $\rightarrow$ action & Rate & All repair/harm & Routed repair/harm & Net gain \\
    \midrule
    Qwen-8B $\rightarrow$ Qwen-32B/8F & 30\% & 26.0/9.9 & 39.2/9.0 & $+9.03\pp{}$ \\
    Qwen-4B $\rightarrow$ Qwen-32B/8F & 30\% & 22.0/8.7 & 32.5/9.0 & $+7.04\pp{}$ \\
    IVL-8B $\rightarrow$ Qwen-32B/8F  & 25\% & 29.2/15.0 & 40.6/13.8 & $+6.68\pp{}$ \\
    IVL-8B $\rightarrow$ IVL-8B/16F   & 30\% & 11.0/13.5 & 18.7/12.7 & $+1.81\pp{}$ \\
    \bottomrule
  \end{tabular}
\end{table}

The decomposition makes the action dependence explicit.  Qwen-32B/8F offers a
large repair pool but also harms 8.7--15.0\% of initially correct answers when
applied indiscriminately.  InternVL/16F is globally harmful because its harm
rate exceeds its repair rate (13.5\% versus 11.0\%), yet the learned policy
enriches repairs in the routed subset (18.7\% versus 12.7\%) and obtains a
positive net gain.  Error probability alone cannot represent this distinction.

\end{document}